\documentclass[runningheads]{llncs}

\usepackage[T1]{fontenc}

\usepackage{graphicx}
\usepackage{multirow}
\usepackage{amsmath}
\usepackage{amssymb}
\usepackage{amsfonts}
\usepackage{bbold}

\usepackage{subcaption}
\usepackage{mathtools}
\usepackage{color}

\usepackage[colorlinks=true,linkcolor=blue,citecolor=blue,urlcolor=blue]{hyperref}
\usepackage{breakurl}
\usepackage[misc]{ifsym}
\usepackage{pifont}
\newcommand{\cmark}{\ding{51}}%
\newcommand{\xmark}{\ding{55}}%

\def\etal{\textit{et al}. }

\def\vs{\textit{v.s. }}

\begin{document}
%
\title{Radiomics-Informed Deep Learning for Classification of Atrial Fibrillation Sub-Types from Left-Atrium CT Volumes }

\titlerunning{Radiomics-Informed Deep Learning (RIDL)}
%
\author{Weihang Dai\inst{1}\orcidID{0000-0002-8619-236X}, Xiaomeng Li\inst{1(}\textsuperscript{\Letter}$^)$\orcidID{0000-0003-1105-8083}, Taihui Yu\inst{2,3}, Di Zhao\inst{4}, Jun Shen\inst{2,3}, Kwang-Ting Cheng\inst{1}\orcidID{0000-0002-3885-4912}}

\institute{The Hong Kong University of Science and Technology \\
\email{eexmli@ust.hk}\\
\and
Sun Yat-Sen University \\
\and
Sun Yat-Sen Memorial Hospital\\
\and
Chinese Academy of Sciences
}

%

%
\maketitle              
\begin{abstract}

Atrial Fibrillation (AF) is characterized by rapid, irregular heartbeats, and can lead to fatal complications such as heart failure. 
The disease is divided into two sub-types based on severity, which can be automatically classified through CT volumes for disease screening of severe cases.  
However, existing classification approaches rely on generic radiomic features that may not be optimal for the task,
whilst deep learning methods tend to over-fit to the high-dimensional volume inputs.  
In this work, we propose a novel radiomics-informed deep-learning method, RIDL, that combines the advantages of deep learning and radiomic approaches to improve AF sub-type classification. 
Unlike existing hybrid techniques that mostly rely on na\"ive feature concatenation, we observe that radiomic feature selection methods can serve as an information prior, and propose supplementing low-level deep neural network (DNN) features with locally computed radiomic features. This reduces DNN over-fitting and allows local variations between radiomic features to be better captured. 
Furthermore, we ensure complementary information is learned by deep and radiomic features by designing a novel feature de-correlation loss. Combined, our method addresses the limitations of deep learning and radiomic approaches and outperforms state-of-the-art radiomic, deep learning, and hybrid approaches, achieving 86.9\% AUC for the AF sub-type classification task. Code is available at \href{https://github.com/xmed-lab/RIDL}{https://github.com/xmed-lab/RIDL}.

\end{abstract}

\section{Introduction}

Atrial fibrillation (AF) is a cardiac disease characterized by rapid, irregular heartbeats \cite{go2001prevalence}. The disease can lead to stroke and heart failure, and has a mortality rate of almost 20\% \cite{pastori2015incidence,gomez2016causes,lee2017atrial}. AF is classified as either persistent atrial fibrillation (PeAF), where abnormal heart rhythms occur continuously for more than seven days, or paroxysmal atrial fibrillation (PaAF), where the heart rhythm returns to normal within seven days. Although AF can be treated through a procedure called catheter ablation, PeAF cases have high recurrence rates and often require re-intervention \cite{january20142014}. 
Accurate knowledge of the disease type is therefore highly valuable for treatment planning and has high prognostic value \cite{yang2022development}.

Clinical studies have discovered a strong relationship between AF and epicardial adipose tissue (EAT), a fat depot layer on the surface of the myocardium that can cause inflammation and disrupt cardiac function \cite{shamloo2019epicardial,gaeta2017epicardial}. 
Recent works have shown that automatic classification of AF sub-types can be done using CT volumes of the left atrium and surrounding EAT, which can be used to screen for patients with high risk of PeAF. 
Huber \etal \cite{huber2022relationship} showed that EAT volume, approximated from left-atrium CT images, can be used as a predictor for AF recurrence. 
Yang \etal \cite{yang2022development} trained a random forest model to classify AF sub-type based on radiomic features and volume measurements, achieving 85.3\% AUC. Although these methods demonstrate the usefulness of radiomic features for AF sub-type classification, such features are generic and not specific to the task, which can limit model performance \cite{li2017fully}. Radiomic features also rely on summary statistics such as entropy or homogeneity to obtain global descriptors, and these have limited effectiveness when capturing local feature variations \cite{sun2020deep}.

Deep learning has achieved outstanding results on medical imaging analysis tasks, largely due to its ability to learn task-specific features and complex relations between them \cite{truhn2019radiomic}. 
Na\"ively using deep neural networks (DNNs) to predict AF sub-types from CT volumes yields poor results however due to over-fitting on high-dimensional volume inputs (see results for DNN in Table 1). 
Existing works have attempted to combine deep and radiomic features through methods such as direct concatenation \cite{cui2022ct,wang2021deep}, attention modules \cite{saeed2022tmss}, or contrastive learning between feature types \cite{zhao2021unsupervised}. Although these methods propose different ways of using both approaches, \textit{they do not explicitly address the limitations of either approach or explore ways to combine their complementary advantages}. 

\begin{figure}%
\centering
\includegraphics[width=0.70 \columnwidth]{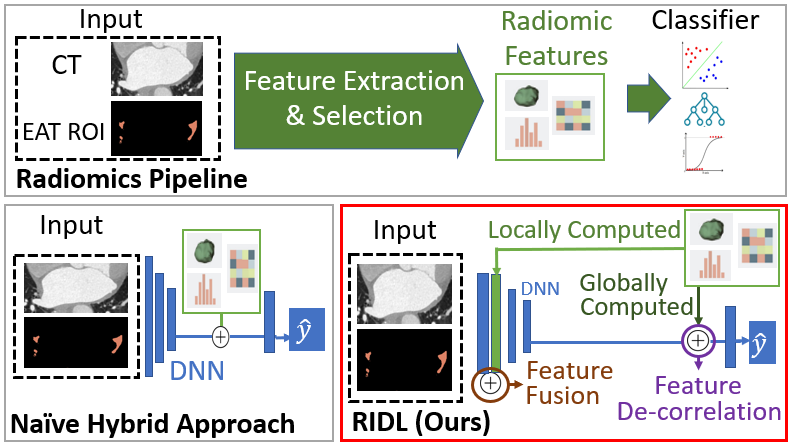}%
\caption{Overview of our method compared with alternative approaches. Radiomic modeling uses feature-selection algorithms to select predictive features. Na\"ive hybrid approaches directly concatenates these radiomic with deep features. Our method also fuses \textit{locally computed} radiomic features with low-level DNN features and encourages \textit{complementary} deep and radiomic features to be learned. }%

\label{detail}
\end{figure}

In this work, we propose a novel approach to atrial fibrillation sub-type classification from CT volumes by integrating radiomic and deep learning methods. 
We note that textural radiomic features identified by feature selection methods \textit{can serve as an information prior} to supplement low-level features from DNNs, since they are designed to capture low-level context and have predictive power \cite{zhang2022deep}. To this end, we \textit{locally} calculate radiomic features based on patches surrounding each voxel, and perform feature fusion with low-level DNN features. This provides the DNN with pre-defined features known to be relevant to the task to reduce over-fitting, and also allows spatial relations between radiomic features to be learned. 
Furthermore, we encourage the DNN to learn features \textit{complementary} to radiomic features to obtain more comprehensive signals and design a novel feature de-correlation loss. The overall framework, which we term \textbf{R}adiomics-\textbf{I}nformed \textbf{D}eep \textbf{L}earning (RIDL), is illustrated in Fig. 1.
Unlike existing works, our method is designed to directly addresses the limitations of both deep learning and radiomic approaches and achieves state-of-the-art performance on AF sub-type classification. To summarize our key contributions:

\begin{itemize}

\item
We propose a novel radiomics-informed deep learning (RIDL) method for AF sub-type classification from CT volumes, which achieves state-of-the-art results and can be used to screen for patients with high risk of PeAF.

\item Our method uses a novel approach of fusing \textit{locally computed} radiomic features with low-level DNN features to improve capturing of local context.

\item
Furthermore, we enforce feature de-correlation using a novel feature-bank design to ensure \textit{complementary} deep and radiomic features are extracted.

\end{itemize}







\section{Methodology}

We combine radiomic and deep learning approaches using two novel components: 1) feature fusion of local radiomic features and low-level DNN features to improve local context, 2) encouraging complementary deep and radiomic features through feature de-correlation. These are illustrated in Fig. 2 and explained in detail below. Our dataset $\mathcal{D} \coloneqq \left \{ (x_i, y_i) \right \}_{i=1}^{N}$ includes $N$ samples of input $x_i$ and binary label $y_i$, where 0 indicates PaAF and 1 indicates PeAF. $x_i$ has two channels, one consisting of the 3D CT volume centered around the left atrium and the other the binary region-of-interest (ROI) mask indicating EAT. The ROI is obtained through Hounsfield value thresholding such that all voxels valued between -250 and 0 are identified as EAT \cite{huber2022relationship,yang2022development}.

\begin{figure}%
\centering
\begin{subfigure}{0.5\columnwidth}
\centering
\includegraphics[width=0.98 \columnwidth]{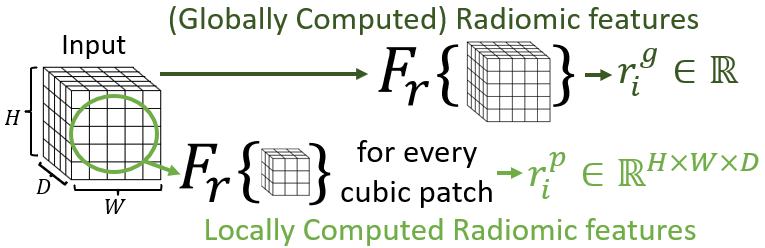}%
\caption{}%
\label{detail_a}%
\end{subfigure}\hfill%
\begin{subfigure}{0.46\columnwidth}
\centering
\includegraphics[width=0.98 \columnwidth]{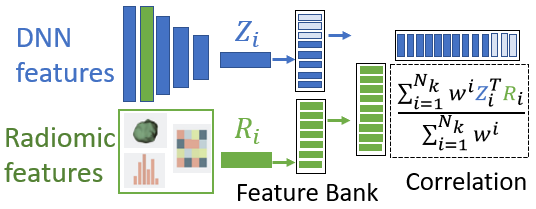}%
\caption{}%
\label{detail_b}%
\end{subfigure}\hfill%

\caption{Illustration of the two main components of our method. (a) Calculation of global \vs local radiomic features. (b) Feature bank implementation for feature de-correlation. Light blue features indicate features in current training iteration}
\label{detail}
\end{figure}

\subsection{Feature fusion of locally computed radiomic features with low-level DNN features for improved local context}

Under the radiomics pipeline, a large set of features, typically more than a thousand, is first extracted by performing calculations over the volume and ROI input $x_i$.
Feature selection methodologies such as mutual information (MI), principal component analysis (PCA), or LASSO regularization, are then used to identify predictive features for classification \cite{zhang2022deep}. 
Radiomic features are classified into shape, first-order statistical features, and texture features. Texture features are designed to capture local variations and use measures such as Gray-level Co-occurrence Matrices (GLCM) to reflect second-order textural distributions. 
Conventional statistics such as entropy and correlation are then used to summarize these measures \cite{zwanenburg2020image}, but these tend to be limited in their ability to capture local heterogeneity, such as the varying textures on the surface of a cancer tumor. Although DNN's are more effective at capturing local variations, they can overfit without sufficient data for training \cite{truhn2019radiomic}.

Unlike existing works that na\"ively concatenate radiomic and deep features before the classification layer \cite{cui2022ct,wang2021deep}, we observe that textural features selected through radiomics feature selection algorithms are known to be predictive and \textit{can be used as prior knowledge} to improve low-level DNN features. 
Given radiomic feature extractor $F_r$, the global radiomic feature, $r_i^{g} \in \mathcal{R}$, for input $x_i$ is represented by:
\begin{equation}
    r_i^{g} = F_r(x_i) \:.
\end{equation}
Our method applies feature calculations \textit{locally} to cubic patches centered around each voxel, such that features are obtained on a voxel basis and reflect the statistics of the neighbouring region. For a cubic patch with radius $p$ and input $x_i$, the local feature at location $(h,w,d)$, denoted by $r^{p}_{i,(h,w,d)}$, is obtained by performing $R$ on the cubic patch in $x_i$ centered around $(h,w,d)$:
 \begin{equation}
    r^{p}_{i,(h,w,d)} = F_r(x_{i,[h-p:h+p, w-p:w+p, d-p:d+p]}) \:,
\end{equation}
where the input of $F_r$ is the cubic sub-volume. This process is illustrated in Fig. 2a. 
Local features can be calculated for multiple texture features and patch size $p$, which are then concatenated to obtain $r^{l}_i \in \mathcal{R}^{L \times H \times W \times D} $, where $L$ is the total number of features used and $H$, $W$, and $D$ are original input dimensions. We note that only texture radiomic features are used for local calculation since they are specifically intended to capture local context.

$r^{l}_i$ is then concatenated with low-level DNN features, $z_i \in \mathcal{R}^{C \times H \times W \times D}$, to supplement the DNN with local radiomic features. To effectively fuse the features, we apply a channel attention module, $\mathcal{A}$, following the design in \cite{woo2018cbam}: 

\begin{equation}
    z_i'=\mathcal{A}(r^{l}_i \oplus z_i) \otimes (r^{l}_i \oplus z_i)
\end{equation}
where $z_i'$ is the fused feature, $\oplus$ is channel concatenation, and $\otimes$ is element-wise multiplication. The learned attention tensor $\mathcal{A}(r^{l}_i \oplus z_i)$ has dimensions $(C+L)\times 1 \times 1\times 1$ and is broadcasted along the volume dimension, such that attention is applied channel-wise and spatial feature distributions are preserved.




\subsection{Encouraging complementary deep and radiomic features through feature de-correlation}

Global radiomic features are also included in our model by concatenation with high-level DNN features before the classification layer. 
Unlike existing approaches however, we encourage our DNN to learn features \textit{complementary} to radiomic features by enforcing de-correlation between the two. 
This ensures that different variations are captured, which provides a more comprehensive signal to the classification layer. 

Accurate approximation of correlation requires large batches sizes however, which requires large GPU memory and can affect model convergence \cite{keskar2016large}. 
We instead propose a novel feature-bank implementation with exponential weighting to estimate sample statistics. 
Every iteration, we save DNN and global radiomic features in feature-bank $K$, which holds up to $N_{k}$ features in a first-in first-out queue. After a warm-up period, we calculate the sample correlation
using an exponential weighting scheme. 
Given weight parameter $w < 1$, and the normalized deep feature $Z_i$ and radiomic feature $R_i$ from $K$, we calculate feature de-correlation loss $\mathcal{L}_{corr}$ as:
\begin{equation}
\mathcal{L}_{corr} = || \frac{\sum_{i=1}^{N_{k}} w^i Z_{i}^T R_{i}}{\sum_{i=1}^{N_{k}} w^i} ||_1 \:.
\end{equation}
The first $B$ samples, where $B$ is the batch size, belong to the training sample of the current iteration, and their losses are back-propagated to encourage deep features to have zero correlation with radiomic features. This process is illustrated in Fig. 2b.  Although feature banks have been used in techniques such as contrastive learning to address batch size limitations \cite{he2020momentum,wu2018unsupervised}, we are the first to formulate this technique \textit{for feature de-correlation}.

\subsection{Overall framework}

The DNN model uses raw CT volumes concatenated with ROI masks as input. 
Global and local radiomic features are pre-computed for input into the feature layer. Binary cross-entropy is used for AF sub-type classification loss $\mathcal{L}_{cls}$:
\begin{equation}
    \mathcal{L}_{cls} = \frac{1}{N} \sum_{i=1}^N[y_i \ln(\hat{y}_i) + (1-y_i) \ln(1-\hat{y}_i)]
\end{equation}
where $\hat{y}_i$ is the model prediction for sample $y_i$. The model is trained together with feature de-correlation loss $\mathcal{L}_{corr}$ and its loss weighting, $w_{corr}$. To provide further regularization and prevent over-fitting, we perform an additional self-reconstruction task, using loss $\mathcal{L}_{rec}$, which we describe in more detail in the supplementary materials.
The overall loss function is then:
\begin{equation}
    \mathcal{L} = \mathcal{L}_{cls} + w_{corr} \mathcal{L}_{corr}  + \mathcal{L}_{rec} \:.
\end{equation}

\section{Experiments}

\subsection{Implementation Details}
\subsubsection{Dataset}
We use a dataset of 172 patients containing 94 PaAF and 78 PeAF cases collected from the Sun Yat-Sen Memorial Hospital in China. CT volumes are centered on the left atrium and normalized to between -1 and 1. ROI masks for EAT are obtained through Hounsfield value thresholding between -250 and 0. Volumes are resized to the same aspect ratio to ensure consistent dimensions across samples. We use an input size of 96x128x128 voxels and apply zero padding for smaller volumes. 
We use five-fold cross-validation and report average test performance across folds. 
Cross-validation is implemented by splitting the dataset into five equal subsets and using three subsets for training, one subset for validation, and one subset for testing. A rolling scheme is used such that different validation and test subsets are used for each of the five folds. 
Data acquisition procedures and statistics are given in the supplementary materials. 

\subsubsection{Setup}
We use the PyRadiomic package \cite{van2017computational} to extract radiomic features from the input volumes and masks. Using the cross-validation splits, we perform feature selection and classification using LASSO regularized logistic regression.
LASSO regularization consistently selects four radiomic features as the ones with the most significant predictive power: maximum 3D diameter, Maximum 2D Diameter, Maximum voxel value, and normalized inverse difference of GLCM (glcm\_Idn). The texture feature glcm\_Idn is calculated locally for $p\in \{1, 2, 5,10\}$ to obtain local radiomic features $r_i^l \in  \mathcal{R}^{4 \times 96 \times 128 \times 128}$. 

For our DNN network, we use a modified 3D U-Net \cite{cciccek20163d} (abbreviated as m3DUNet) with skip connections between the encoder and decoder removed to enhance bottle-neck feature compression. Bottle-neck features are averaged across spatial dimensions for classification, whilst decoder outputs are used for self-reconstruction regularization. The model is trained using the Adam optimizer with learning rate $10^{-4}$ for 100 epochs and 0.1 decay at 30 epochs. We use batch size $B=1$, feature bank size $N_k = 25$, and warm-up period of one epoch. We use $w_{corr}=2$ for de-correlation loss weighting, which was chosen based on the validation splits. Mean and standard deviation of ten runs are reported. Additional experiments and details are included in the supplementary materials.

\subsection{Comparison with State-of-the-art Methodologies}

We compare our method with alternative state-of-the-art approaches based on radiomics, deep learning, and hybrid techniques. 
Deep and hybrid volume-based classification methods \cite{saeed2022tmss,zhao2021unsupervised,lee2022moving} are adapted to our task since there are no existing works for AF sub-type classification. We use the same encoder for all deep architectures for fair comparison, except for methods that are architecture specific. A na\"ive feature concatenation method is used as our baseline for the hybrid approach. Radiomic features for the hybrid approach are selected through LASSO regularization as it is the most effective. Results are shown in Table~1. 

\begin{table}
\centering
\caption{Comparison with state-of-the-art methods for radiomic, deep learning, and hybrid approaches. Selector$^{\ast}$ refers to the feature selection method. Hybrid$^{\#}$ methods use radiomic features selected by LASSO regularization, which is the most effective.  DNN$^{\star}$ is a na\"ive implementation using the m3DUNet model. Baseline$^{\dagger}$ is a na\"ive hybrid implementation using simple feature concatenation.
}
\label{tab:sota}
\begin{tabular}{l|l|l|c|c|c|c}


\hline
Type  & Selector$^{\ast}$  & Classifier & AUC (\%) & MAP (\%) & F1 (\%) & Acc. (\%)\\

\hline
\multirow{4}{*}{\textit{Radiomic}} & \multirow{3}{*}{{\begin{tabular}[c]{@{}l@{}}Mutual\\Information\end{tabular}}} & SVM &  74.2 &  65.2 & \textbf{71.1} & \textbf{74.7}\\
 & & Random Forest \cite{yang2022development} &  72.4 & 63.4 & 70.5 & 72.4 \\

& & Logistic Regression &  68.7 & 59.1 & 65.7 & 69.3 \\


\cline{2-7}
&  \multirow{1}{*}{LASSO} & Logistic Regression   &\textbf{83.4} & \textbf{81.7} & 69.1 & 73.9 \\
\hline

\end{tabular}

\vspace{-0.6em}

\begin{tabular}{l|l|l|c|c|c|c}

\multicolumn{6}{l}{\textit{ }}\\ 
\hline
Type & Method  & Model & AUC (\%) & MAP (\%) & F1 (\%) & Acc. (\%)\\
\hline
\multirow{2}{*}{\begin{tabular}[c]{@{}c@{}}\textit{Deep}\\\textit{Learning}\end{tabular}} 
 & Lee \etal \cite{lee2022moving} & f-rMC5  &  63.3 $\pm$ 6.3 & 64.8 $\pm$ 3.9 & 31.5 $\pm$ 7.6 & 63.3 $\pm$ 4.3 \\
 &  DNN$^{\star}$ & m3DUNet & \textbf{77.2 $\pm$ 1.5} & \textbf{73.7 $\pm$ 1.4} & \textbf{68.4 $\pm$ 1.4} & \textbf{70.7 $\pm$ 1.8} \\

\hline
\end{tabular}

\vspace{-0.6em}







\begin{tabular}{l|l|l|c|c|c|c}

\multicolumn{7}{l}{\textit{ }}\\ 
\hline
Type & Method    & Model & AUC (\%) & MAP (\%) & F1 (\%) & Acc. (\%)\\
\hline
\multirow{4}{*}{\textit{Hybrid$^{\#}$}} & Zhao \etal \cite{zhao2021unsupervised}  &  m3DUNet &  85.4 $\pm$ 0.1 & 85.2 $\pm$ 0.3 & 70.6 $\pm$ 1.2 & 73.5 $\pm$ 1.0 \\

& TMSS \cite{saeed2022tmss}  &  TMSS &  84.0 $\pm$ 0.5 & 82.3 $\pm$ 0.2 & 71.6 $\pm$ 3.2 & 74.5 $\pm$ 1.2 \\

& Baseline$^{\dagger}$   & m3DUNet &  85.8 $\pm$ 0.5 & 84.5 $\pm$ 0.6 & 73.9 $\pm$ 1.6 & 75.7 $\pm$ 1.4 \\

& RIDL (ours)  & m3DUNet  & \textbf{86.9 $\pm$ 0.6} & \textbf{86.3 $\pm$ 0.6} & \textbf{74.7 $\pm$ 1.5} & \textbf{76.9 $\pm$ 1.0} \\
\hline

\end{tabular}

\end{table}

We can see that hybrid methods outperform radiomic and deep methods in general. Our method, RIDL, achieves the best results across all metrics however and improves AUC by 1.1\% over the baseline method (86.9\% \vs 85.8\%) 
and 3.5\% over the best radiomics approach (86.9\% \vs 83.4\%). 

\subsection{Ablation}

\subsubsection{Component analysis}

We perform ablation experiments to demonstrate improvements from using local radiomic features, global radiomic features, and feature de-correlation loss. Results are shown in Table 2.

 \begin{table}
\fontsize{9pt}{10.8pt}\selectfont
\begin{center}
\caption{Ablation study for proposed method. 
$\checkmark$ indicates use of global radiomic features ($R_i^g$ ), local radiomic features ($r_i^l$) and feature de-correlation loss ($\mathcal{L}_{corr}$). DNN$^{\star}$ is a na\"ive implementation using the m3DUNet model. Baseline$^{\dagger}$ is a na\"ive hybrid implementation using simple feature concatenation.
}
\label{table:abl_age}
\begin{tabular}{l|ccc|c|c}
\hline
Method & $R_i^g$ & $r_i^l$ & $\mathcal{L}_{corr}$  & AUC (\%) & MAP (\%) \\
\hline

DNN$^{\star}$           & &  &  & 77.2 $\pm$ 1.5 & 73.7 $\pm$ 1.4 \\
DNN$^{\star}$ + Local Radiomic Features          &   & \cmark &  &  78.8 $\pm$ 1.1 & 74.9 $\pm$ 1.1 \\
Baseline$^{\dagger}$           &  \cmark &  &  & 85.8 $\pm$ 0.5 & 84.5 $\pm$ 0.6 \\
Baseline$^{\dagger}$ + Local Radiomic Features              & \cmark &  \cmark &  & 86.3 $\pm$ 0.9 & 84.9 $\pm$ 8.0 \\
RIDL (ours) & \cmark &  \cmark & \cmark &  \textbf{86.9 $\pm$ 0.6} & \textbf{86.3 $\pm$ 0.6} \\

\hline
\end{tabular}
\end{center}
\end{table}

We can see that including local radiomic features,  improving AUC by up to 1.6\% when included with a standard DNN (78.8\% \vs 77.2\%). Using feature de-correlation further boosts performance and leads to the best overall results.

\subsubsection{Effectiveness of radiomic feature selection }

To demonstrate the effectiveness of radiomic feature selection as prior knowledge for feature fusion, we compare with results from using features discarded by radiomics feature selection. We randomly select three discarded features to generate local features $r_i^l$ as input whilst keeping other components constant. Results are shown in Table~3.

 \begin{table}
\fontsize{9pt}{10.8pt}\selectfont
\begin{center}
\caption{Results using different texture radiomic features for input $r_i^l$ as displayed by their PyRadiomics key \cite{van2017computational}. "Selected" indicates whether the feature was selected or discarded by the radiomic feature selection algorithm. 
}
\label{table:abl_age}
\begin{tabular}{l|c|c|c}
\hline
Feature used for local calculation & Selected & AUC (\%) & MAP (\%) \\
\hline

gldm\_DependenceNonUniformityNormalized  & \xmark & 86.1 $\pm$ 0.8 & 85.5 $\pm$ 0.9 \\
glrlm\_LongRunEmphasis    & \xmark  & 86.4 $\pm$ 0.8 & 85.4 $\pm$ 0.8 \\ 
gldm\_LargeDependenceEmphasis   & \xmark & 85.6 $\pm$ 0.7 & 84.6 $\pm$ 0.9 \\ 

glcm\_IDN (ours) & \cmark & \textbf{86.9 $\pm$ 0.6} & \textbf{86.3 $\pm$ 0.6} \\
\hline
\end{tabular}
\end{center}
\end{table}

We can see that using discarded features leads to worse performance in general. Given the large set of radiomic features, it is possible some discarded features may outperform selected features due to differences in global and local computation. 
Nevertheless, our results indicate that the radiomic feature selection process serves as an reasonable information prior. Our work is the first to propose \textit{fusing locally computed radiomic features with low-level DNN features}, and we leave detailed local feature selection methods to future works.




\section{Conclusion}

In this work, we propose a new approach to atrial fibrillation sub-type classification from CT volumes by integrating radiomic and deep learning approaches through a radiomics-informed deep learning method, RIDL. Our method is based on two key ideas: feature fusion of locally computed radiomic features with low-level DNN features to improve local context, and encouraging complementary deep and radiomic features through feature de-correlation. Unlike existing hybrid approaches, our method specifically addresses the advantages and limitations of both techniques to improve feature extraction. 
We achieve state-of-the-art results on AF sub-type classification and outperform existing radiomic, deep learning, and hybrid methods.

Future improvements to RIDL can be made by introducing more sophisticated local radiomic features selection methods, given the large set features to choose from. Experiments on larger datasets or alternative tasks can also be done to provide more empirical support, since current results show only slight improvements over baseline. 
These issues may be addressed in future works. 
%
Overall, our method is a novel way of combining radiomic and deep learning approaches, and can be used to improve accuracy of PeAF screening from CT volumes for better preventive care of high-risk patients. 

\section*{Acknowledgement}
This work was supported in part by grants from Hong
Kong Innovation and Technology Commission (Project no.
ITS/030/21 \& Project no. PRP/041/22FX), and by Foshan HKUST Projects under FSUST21-HKUST10E and FSUST21-HKUST11E.

\bibliographystyle{splncs04}
\bibliography{paper1244_bib}
%




\end{document}


%
\title{Radiomics-Informed Deep Learning (RIDL)
}

%

\author{Paper ID: 1244}
\institute{}
%

%
\maketitle              

\section{Dataset}
\vspace{-8mm}

\begin{figure}%
\centering

\begin{subfigure}{0.25 \columnwidth}
\centering
\includegraphics[width=1 \columnwidth]{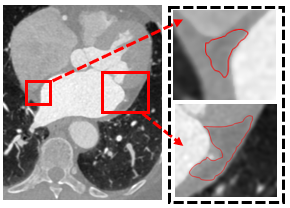}%
\vspace{-2mm}
\caption{}%
\label{featsim_bad}%
\end{subfigure}\hfill%
\begin{subfigure}{0.73 \columnwidth}
\centering
\includegraphics[width=1 \columnwidth]{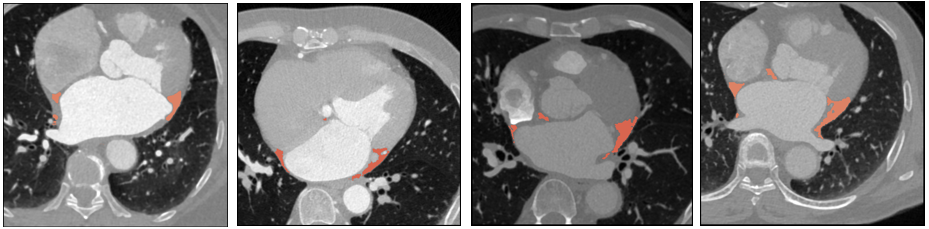}%
\vspace{-2mm}
\caption{}%
\label{featsim_good}%
\end{subfigure}\hfill%

\vspace{-3mm}
\caption{(a) Example of Hounsfield Unit thresholding for Epicardial adipose tissue (EAT) region of interests (ROI) (b)
 Raw CT slices and ROI masks (red) from dataset. CT volumes were captured using a Somatom Force scanner (Siemens Heathineers) centered on the left atrium.
The injection protocol included an injection of 50 $\sim$ 60 mL contrast agent at the rate of 4.0-6.0 mL/s. Tube voltage and tube current was determined automatically according to CARE Dose4D and CARE kV technology. Dimensions of each slice is set to 512 $\times$ 512 pixels.
}
\label{featsim}
\end{figure}

\vspace{-12mm}

\begin{figure}%
\centering
\begin{subfigure}{0.33 \columnwidth}
\centering
\includegraphics[width=0.75 \columnwidth]{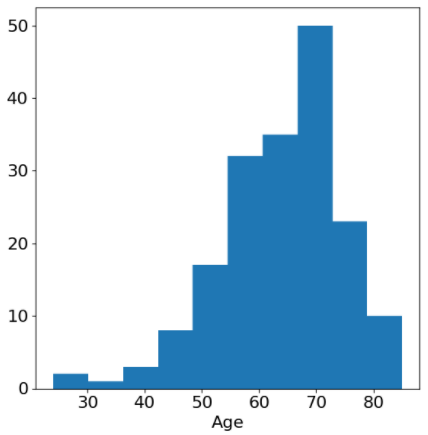}%
\vspace{-2mm}
\caption{Age distribution}%
\label{featsim_bad}%
\end{subfigure}\hfill%
\begin{subfigure}{0.33 \columnwidth}
\centering
\includegraphics[width=0.75 \columnwidth]{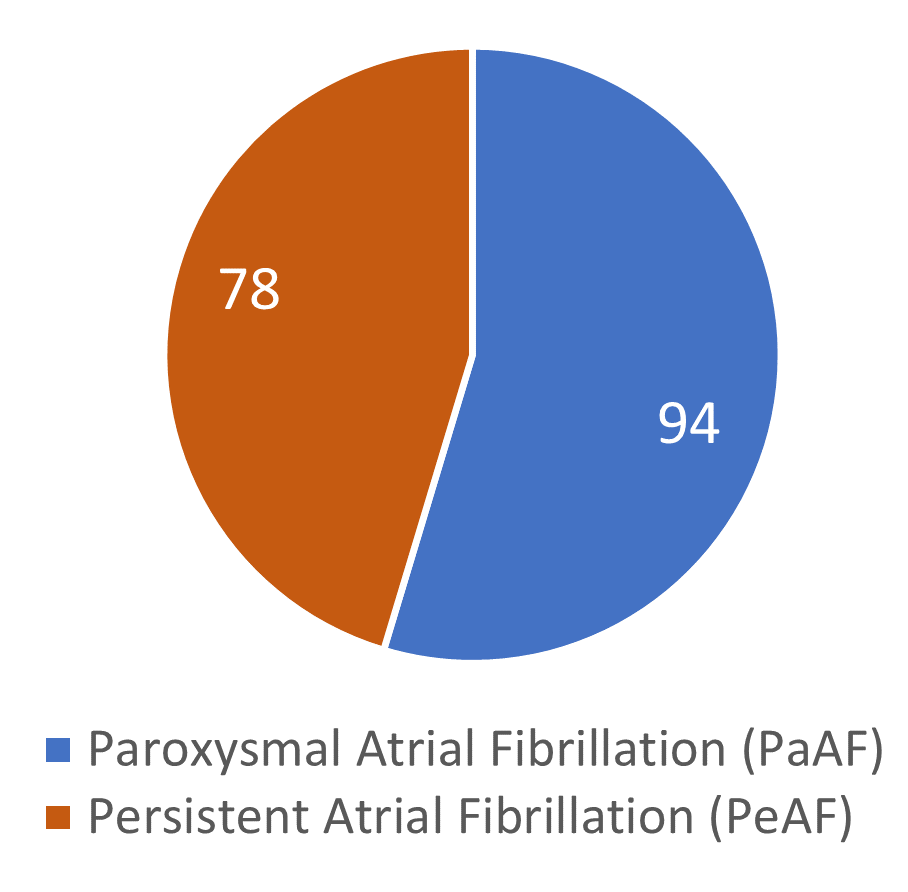}%
\vspace{-2mm}
\caption{Label distribution}%
\label{featsim_bad}%
\end{subfigure}\hfill%
\begin{subfigure}{0.33 \columnwidth}
\centering
\includegraphics[width=0.75 \columnwidth]{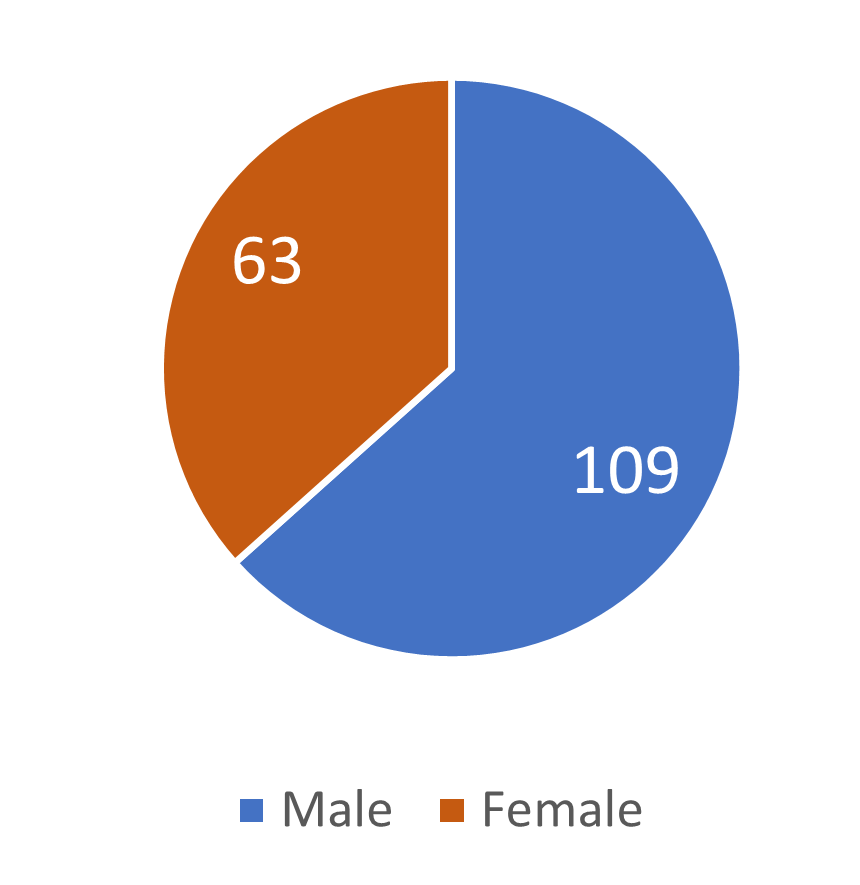}%
\vspace{-2mm}
\caption{Gender distribution}%
\label{featsim_good}%
\end{subfigure}\hfill%

\vspace{-3mm}
\caption{Patient distribution for dataset
}
\label{featsim}
\end{figure}

\vspace{-10mm}

\section{Implementation Details}
\vspace{-12mm}

\begin{table}[hbt!]
\centering
\caption{Additional implementation details for section 3.2}
\label{tab:ablation_param}
\begin{tabular}{|l|l|}
\hline
\textbf{Method}  & \textbf{Description} \\
\hline
Lee \etal [11]         & We directly use their implementation with our dataset and labels \\
\hline
DNN          & We use the m3DUNet architecture without any radiomic features \\
\hline
Zhao \etal [24]          & {\begin{tabular}[c]{@{}l@{}}We pre-train the encoder of m3DUNet following their method and \\  use na\"ive concatenation with radiomic features similar to Baseline. \end{tabular} \\
\hline
TMSS [14]         & {\begin{tabular}[c]{@{}l@{}}We use the EAT ROI mask for segmentation and replace clinical  \\ data with radiomic features. Other components are kept the same. \end{tabular}}

 \\
\hline
Baseline          & {\begin{tabular}[c]{@{}l@{}}We use na\"ive concatenation with radiomic features. Weights are  \\ initialized with models from DNN method to improve convergence. \end{tabular}} \\
\hline
RIDL (ours)          & Same as Baseline, with local radiomic features and de-correlation loss\\
\hline

\end{tabular}

\end{table}


\begin{figure}%
\centering
\includegraphics[width=0.85 \columnwidth]{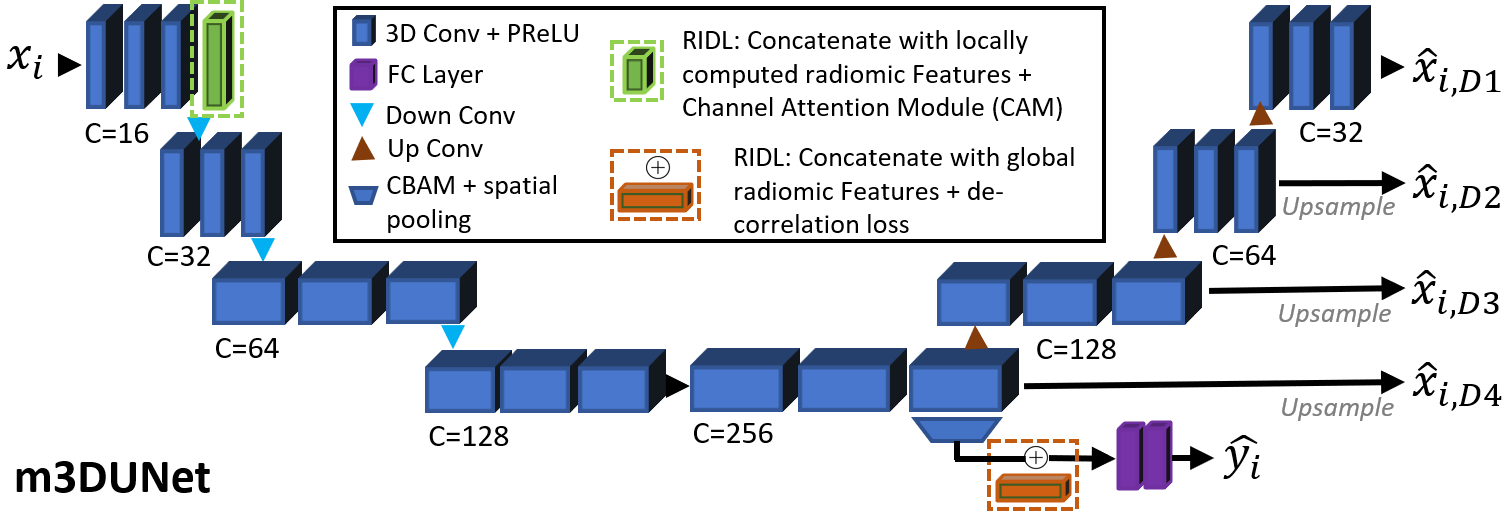}
\caption{Architecture of the modified 3D-Unet (m3DUnet). Residual connections between encoders and decoders of the standard 3D-UNet are removed for feature compression through the bottleneck. Bottleneck features pass through a CBAM [20] attention module to attend to relevant features before spatial pooling and classification with two fully connected layers to obtain $\hat{y}_i$. RIDL supplements low-level deep features (green) and high-level deep features (orange) with radiomic features. Decoder outputs are used for reconstruction regularization. 
}
\vspace{-3mm}
\label{featsim}
\end{figure}

\noindent S-Eq. 1: Calculation of self-reconstruction loss $\mathcal{L}_{rec}$. Input $x_{i}$ consists of a CT channel $c_{i}$, and binary ROI mask $b_{i}$. Reconstruction output $\hat{x}_{i,d}$ consists of $\hat{c}_{i,d}$, and $\hat{b}_{i,d}$ for decoders $d \in \{D1,..,D4\}$.
We use mean absolute error for CT and binary cross-entropy for ROI reconstruction. Loss for decoder $d$, $\mathcal{L}_{rec,d}$, is:
\vspace{-2mm}
\begin{equation}
\begin{split}
    \mathcal{L}_{rec,d} = & \frac{1}{N\times |V|} \sum_{i=1}^N \sum_{v\in V}|\hat{c}_{i,d}(v) - c_{i}(v)| \: + \\
     & \frac{1}{N\times |V|} \sum_{i=1}^N  \sum_{v\in V}[b_{i}(v) \ln(\hat{b}_{i,d}(v)) + (1-b_{i}(v)) \ln(1-\hat{b}_{i,d}(v))]
\end{split}
\end{equation}
where $V$ is the set of all voxels and $(v)$ is used to  indicate value at voxel $v$. 
The overall reconstruction loss follows the training scheme of 3D-UNet:
\begin{equation}
\mathcal{L}_{rec} = \mathcal{L}_{rec, D1}  + \alpha \times (\mathcal{L}_{rec, D2} + \mathcal{L}_{rec, D3} + \mathcal{L}_{rec, D4})
\end{equation}
where $\alpha$ is initialized as 0.33 and decayed by a factor of 0.8 every 10 epochs.

\vspace{-3mm}
\section{Additional Experiments}
\vspace{-10mm}
\begin{figure}%
\centering
\begin{subfigure}{0.48 \columnwidth}
\centering
\includegraphics[width=1 \columnwidth]{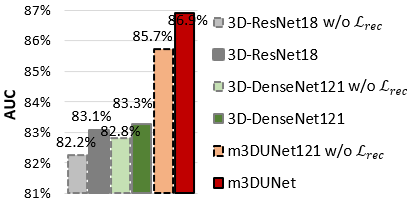}%
\vspace{-2mm}
\caption{}%
\label{featsim_good}%
\end{subfigure}\hfill%
\begin{subfigure}{0.5 \columnwidth}
\centering
\includegraphics[width=0.95 \columnwidth]{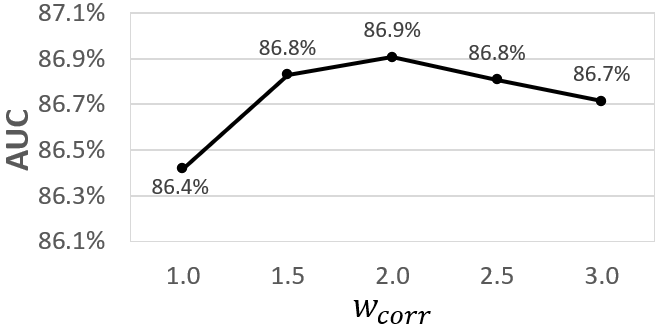}%
\vspace{-2mm}
\caption{}%
\label{featsim_good}%
\end{subfigure}\hfill%

\vspace{-3mm}
\caption{
(a) Results using different encoders. The encoders were directly substituted into m3DUNet. ``w/o $\mathcal{L}_{rec}$" indicates reconstruction regularization was not used. 
(b) Parameter sensitivity of $w_{corr}$. Results stable for $w_{corr} \in (1.5,3)$.\\
}
\label{featsim}
\vspace{-15mm}
\end{figure}